\pdfoutput=1

\documentclass[11pt]{article}

\usepackage[]{ACL2023}

\usepackage{times}
\usepackage{latexsym}

\usepackage[T1]{fontenc}

\usepackage[utf8]{inputenc}

\usepackage{microtype}

\usepackage{inconsolata}

\usepackage{graphicx}
\usepackage[normalem]{ulem}
\useunder{\uline}{\ul}{}
\usepackage{svg}
\usepackage{float}
\usepackage{dblfloatfix}
\usepackage{xurl}
\usepackage{nameref}
\usepackage[german,english]{babel}
\usepackage{multirow}
\usepackage{rotating}

\restylefloat{table}
\title{German Text Embedding Clustering Benchmark}

\author {Silvan Wehrli\textsuperscript{1} \quad Bert Arnrich\textsuperscript{2} \quad Christopher Irrgang\textsuperscript{1} \\ \\
  \textsuperscript{1}Centre for Artificial Intelligence in Public Health Research (ZKI-PH) \\
  Robert Koch Institute, Berlin, Germany \\
  {\tt \{WehrliS,IrrgangC\}@rki.de} \\ \\
  \textsuperscript{2}Digital Health - Connected Healthcare \\
  Hasso Plattner Institute, University of Potsdam, Germany \\
  \tt Bert.Arnrich@hpi.de}

\begin{document}
\maketitle
\begin{abstract}
This work introduces a benchmark assessing the performance of clustering German text embeddings in different domains. This benchmark is driven by the increasing use of clustering neural text embeddings in tasks that require the grouping of texts (such as topic modeling) and the need for German resources in existing benchmarks. We provide an initial analysis for a range of pre-trained mono- and multilingual models evaluated on the outcome of different clustering algorithms. Results include strong performing mono- and multilingual models. Reducing the dimensions of embeddings can further improve clustering. Additionally, we conduct experiments with continued pre-training for German BERT models to estimate the benefits of this additional training. Our experiments suggest that significant performance improvements are possible for short text. All code and datasets are publicly available.
\end{abstract}

\section{Introduction}
Clustering is increasingly used in tasks requiring to group semantically similar text pieces. This includes, for instance, data selection \cite{aharoni-goldberg-2020-unsupervised}, data exploration \cite{voigt-etal-2022-keywordscape}, and neural topic modeling \cite{zhao2021_topic}. One approach for this kind of topic modeling is BERTopic \cite{grootendorst2022bertopic}, which, in principle, uses generic clustering algorithms for text embeddings to find latent topics in text corpora. This is in stark contrast to more traditional topic modeling techniques using Latent Dirichlet Allocation \cite{blei2003lda} or Non-Negative Matrix Factorization \cite{fevotte2011} and representing text as simple bag-of-words. The shift to embedding-based approaches is driven by the continuous development of neural language models, successfully used in natural language understanding (NLU) tasks such as semantic textual similarity \cite{reimers-gurevych-2019-sentence,gao-etal-2021-simcse} or retrieval and reranking \cite{huang2020_retrieval,yates-etal-2021-pretrained}. The availability of plug-and-play frameworks for the computation of vector representation only fosters this trend. One such framework is Sentence Transformers \cite{reimers-gurevych-2019-sentence}, which is used by BERTopic. It provides an extensive collection of pre-trained transformer models and techniques to fine-tune models for similarity-focused language tasks.

Benchmarks help to understand the usefulness of these easily available language models, allowing to compare existing and newly developed models for language tasks of interest. The Massive Text Embedding Benchmark (MTEB, \citealp{muennighoff-etal-2023-mteb}) provides such a benchmark for a wide range of embedding-based tasks (e.g., classification, clustering, or reranking) and datasets from different domains (e.g., online reviews, scientific publications, or social media). MTEB includes a wider range of tasks and focuses on more recent language models than other benchmarks (such as SentEval \cite{conneau-kiela-2018-senteval}). MTEB, offering an easy-to-use API, invites the evaluation of models and submissions to a publicly accessible leaderboard.\footnote{\url{https://huggingface.co/spaces/mteb/leaderboard}}

However, MTEB only considers the English language for the evaluation of clustering. The inclusion of non-English data is important, as the performance of multilingual models may not equal their monolingual counterparts \cite{rust-etal-2021-good}, and as a means to evaluate the potentially strong cross-lingual transfer capability of multilingual models (e.g., \citealp{huang-etal-2019-unicoder}). This work addresses this limitation by providing benchmark datasets and results for German. What is more, MTEB evaluates clustering performance on a single clustering algorithm. This is a suitable approach for such a broad benchmark as it simplifies the evaluation in terms of computational and content-related complexity. From a practical point of view, and specifically for clustering, the evaluation of different algorithms is helpful. Building on the MTEB API, we provide code and evaluation results for a broader range of clustering algorithms. 

Finally, we conduct experiments with continued pre-training. The idea of this additional training is to adapt language models, typically trained on large and heterogeneous data collections, to the data of a specific domain or task, and has been shown to improve performance on downstream tasks (e.g., \citealp{howard-ruder-2018-universal,lee2019-biobert,gururangan-etal-2020-dont}). We analyze the benefit of such adaptive training for clustering within this work.
All code and datasets are publicly available.\footnote{ \url{https://github.com/ClimSocAna/tecb-de}. Additionally, the datasets are officially part of the MTEB library, and results are available in the MTEB leaderboard.}

\begin{table*}[h!]
\resizebox{\textwidth}{!}{
\begin{tabular}{l|lrrrrr}
\textbf{}     & \textbf{}       & \textbf{Unique}  &                                     & \textbf{Size}        & \textbf{Classes}     & \textbf{Avg. chars}   \\
\textbf{Name} & \textbf{Target} & \textbf{Samples} & \multicolumn{1}{r}{\textbf{Splits}} & \textbf{(per split)} & \textbf{(per split)} & \textbf{(per sample)} \\ \hline
BlurbsClusteringS2S   & book titles                                                                       & 17,726                                                            & 28              & 177 to 16,425                                                        & 4 to 93                                                                & 23                                                               \\ \hline
BlurbsClusteringP2P   & \begin{tabular}[c]{@{}l@{}}blurbs\\ (title and blurb)\end{tabular}                & 18,084                                                            & 28              & 177 to 16,425                                                        & 4 to 93                                                                & 664                                                              \\ \hline
TenKGnadClusteringS2S & news article titles                                                               & 10,267                                                            & 9               & 1,436 to 9,962                                                      & 9                                                                      & 51                                                               \\ \hline
TenKGnadClusteringP2P & \begin{tabular}[c]{@{}l@{}}news article texts\\ (title and text)\end{tabular}     & 10,275                                                            & 9               & 1,436 to 9,962                                                      & 9                                                                      & 2,648\\ \hline
RedditClusteringS2S   & submission titles                                                                 & 40,181                                                            & 10              & 9,288 to 26,221                                                     & 10 to 50                                                               & 52                                                               \\ \hline
RedditClusteringP2P   & \begin{tabular}[c]{@{}l@{}}submission descriptions\\ (title and text)\end{tabular} & 40,305                                                            & 10              & 9,288 to 26,221                                                     & 10 to 50                                                               & 902                                                             
\end{tabular}
}
\caption{
Summary of the German benchmark datasets for evaluating the clustering performance of neural language models. Numbers for \textit{Avg. chars} are rounded.
}
\label{tab:datasets}
\end{table*}

\section{Datasets}
\subsection{MTEB Clustering}
\label{subsec:mteb-clustering}
\paragraph{Data sources} The MTEB clustering benchmark covers a range of topical domains and writing styles using data from different sources: arXiv, bioRxic, and medRxiv for scientific publications (e.g., economics or medicine), Reddit for informal social media, Stack Exchange for topical online discussions (such as code), and the 20 Newsgroup dataset \cite{sklearn_api}. 

\paragraph{Text length} MTEB contains two datasets for each data source: A sentence-to-sentence (S2S) dataset compares short texts, and a paragraph-to-paragraph (P2P) dataset compares relatively longer texts. For instance, in the case of arXiv, the S2S dataset only contains publication titles, and the P2P dataset contains the concatenation of titles and abstracts. The two datasets provide models with different amounts of information.

\paragraph{Metric} The evaluation is based on the V-measure \cite{rosenberg-hirschberg-2007-v}. Given a ground truth, the V-measure outputs a score between 0 and 1, measuring homogeneity (clusters contain only one class) and completeness (clusters contain all class samples). MTEB uses topical categories derived from the data, such as the scientific discipline of a publication or newsgroup, as the ground truth.

\paragraph{Data selection} Lastly, datasets in the MTEB clustering benchmark comprise up to 30 random samples of varying size and with a number of different classes drawn from all samples of a data source (splits). 

\subsection{German Additions}
\label{subsec:ger-add}
We follow MTEB's design for German datasets, aiming to simulate a wide range of real-world scenarios by including different domains, text lengths, and clustering complexities (Subsection~\ref{subsec:mteb-clustering}). Compared to English, fewer German open-source datasets seem to exist, which are suitable for this work. Furthermore, some of the open-source datasets in MTEB are generally less relevant for a German benchmark as they contain little to no German content. This includes, for instance, arXiv and Stack Exchange, both mostly English-only data sources.\footnote{There are only around 600 submissions on arXiv (following best practices for search) at the time of writing (\url{https://info.arxiv.org/help/faq/multilang.html}). Similarly, Stack Exchange content is almost exclusively in English: \url{https://meta.stackexchange.com/questions/13676/do-posts-have-to-be-in-english-on-stack-exchange}}
We have identified three openly available German data sources relevant to this benchmark. In the following, we discuss these data sources and the constructed benchmark datasets in more detail (see Table~\ref{tab:datasets} for a summary).\footnote{We provide the scripts used to create these datasets in the Hugging Face repositories linked in our GitHub repository.}

\paragraph{Blurbs} As a first data source, we use data from the GermEval 2019 shared task on hierarchical blurbs classification \cite{remus-etal-2019-germeval}. This data consists of German book metadata, including titles, blurbs (short, promotional descriptions of books), and genres. Even though blurbs are not part of MTEB, the data is well-suited: it is open source and contains topical texts of different lengths (titles, blurbs). What is more, three levels of genres express different levels of detail. The most general genres, for instance, include \textit{Sachbuch} (non-fiction) or \textit{Literatur und Unterhaltung} (literature and entertainment). Secondary and tertiary genres are increasingly specific (e.g., \textit{Fantasy} (fantasy) and \textit{Historische Fantasy} (historical fantasy)). We use this information to evaluate a model's ability to cluster at different granularity (i.e., the ground truth).
We build two datasets, one that only includes book titles (BlurbClusteringS2S) and one that includes the concatentation of titles and blurbs (BlurbsClusteringP2P). The design is based on MTEB's arXiv-based clustering tasks, which use arXiv's two-level categorization (e.g., math and numerical analysis) to simulate cluster granularity.
\\\indent More concretely, we create 10 splits (subsamples) that consider only the broadest category (coarse clustering) and, similarly, 10 splits that consider the second-level genre (fine-grained clustering across all top-level genres). We randomly selected between 10 and 100 percent of the available data for each split. Lastly, we create eight splits by splitting the data based on the top-level genre and considering the second-level genre (fine-grained clustering within a genre).\footnote{We only consider samples with one top-level and up
to two second-level genres. If a sample has two second-level genres, we select the less frequent one (assuming it is more descriptive) to make the label selection less ambiguous.}

\begin{figure}[H]
    \begin{otherlanguage}{german}
        \begin{quote}
            \textit{Der Krieg der Trolle (4)} \\\\
            \textit{Im Land zwischen den Bergen ist die Zeit des Friedens vorbei. Krieg liegt in der Luft, und dann taucht auch noch ein tödlich verwundeter Zwerg im südlichen Hochland von Wlachkis auf – Ereignisse, die wie ein dunkler Schatten auf dem Land liegen. [...]}
        \end{quote}
    \end{otherlanguage}
    \caption*{Example for a book title and blurb from the main category \textit{Literatur \& Unterhaltung} respectively \textit{Fantasy} and \textit{Abenteuer-Fantasy} (second level).}
\end{figure}

\paragraph{News articles} As a second source, we use data from the One Million Posts Corpus \cite{Schabus2017}, inspired by the 10kGNAD dataset\footnote{\url{https://tblock.github.io/10kGNAD}}. 10kGNAD extracts news article information from the One Million Posts Corpus, which consists of annotated user comments (including the corresponding news articles) posted to an Austrian newspaper website. There are nine news categories such as \textit{Wissenschaft} (science) or \textit{Web}, and we use these categories as ground truth for the evaluation. We build two datasets: TenKGnadClusteringS2S, only using article titles, and TenKGnadClusteringP2P dataset, using the whole article texts. We follow MTEB's TwentyNewsgroupsClustering (consisting of news article titles and newsgroups) data selection strategy and draw 10 random samples of varying sizes (selecting at least 10\% of all data).

\begin{figure}[H]
    \begin{otherlanguage}{german}
        \begin{quote}
        \textit{Stoke holt Shaqiri von Inter \\\\
        Arnautovic-Klub zahlt Rekordsumme für Schweizer \\\\
        Stoke-on-Trent/Mailand – Xher\-dan Sha\-qi\-ri wechselt von Inter Mailand zu Stoke City und wird damit Teamkollege von Marko Arnautovic. [...]}
        \end{quote}
    \end{otherlanguage}
    \caption*{Example of a news article consisting of the title, the subheadline and the article text from the \textit{Sport} news section.}
\end{figure}

\paragraph{Reddit} We use data from Reddit as a third data source which we retrieved from the official Reddit API\footnote{\url{https://www.reddit.com/dev/api}}. More precisely, we have collected popular (i.e., hot and top) submissions to 80 German Subreddits such as \textit{r/Bundesliga}, \textit{r/Finanzen}, or \textit{r/reisende}.\footnote{A Subreddit is a topic-specific forum on Reddit, and a submission is a post to a Subreddit. We select active German Subreddit based on desk research and filter for German submissions if Subreddits also contain non-German submissions.} We do not disclose the raw data. Instead, we provide the submission ids and scripts to reproduce the datasets in our GitHub repository. Additionally, Figure~\ref{fig:reddit-bars} in the appendix summarizes the collected data. Our approach is motivated by data privacy and sharing considerations, as discussed in detail in the \nameref{sec:eth-stat}.

In any case, we construct two datasets from the collected data:  SubredditClusteringS2S, which only considers the submission titles, and SubredditClusteringP2P, which combines submission titles and texts. Subreddits We follow the data selection used for MTEB's Reddit-based datasets and build 10 splits with submissions from 10 to 50 randomly selected Subreddits.

\begin{figure}[H]
    \begin{otherlanguage}{german}
        \begin{quote}
        \textit{Wieviel ``Trinkgeld'' für Lieferdienste? \\\\
        Wie viel gebt ihr - sofern ihr Lieferdienste wie Lieferando etc. nutzt - den Fahrern Trinkgeld? Richtet ihr euch nach der 10\% Faustregel in der Gastro?}
        \end{quote}
    \end{otherlanguage}
    \caption*{Example for a submission consisting of the title and text to the German Subbredit \textit{r/Finanzen}.}
\end{figure}

\section{Evaluation Setup}
\subsection{Models}
\label{subsec:eval-models}
We select a range of transformer-based models \citep{NIPS2017_3f5ee243} based on their ability to process German text, architecture, and pre-training methods.\footnote{Table~\ref{tab:hf-model} in the appendix lists the repositories of all models.} For all models, similar to MTEB, we use the mean of a model's output embeddings as text embedding (mean-pooling).

\paragraph{Monolingual models} We include the monolingual GBERT and GELECTRA models \citep{chan-etal-2020-germans}, both based on the BERT \citep{devlin-etal-2019-bert} architecture. GBERT uses whole word masking (WWM) for pre-training\footnote{The originally proposed BERT architecture \cite{devlin-etal-2019-bert} masks subword tokens during pre-training. The authors introduced whole word masking after the publication: \url{https://github.com/google-research/bert/commit/0fce55}.}, while GELECTRA uses the ELECTRA pre-training method \citep{Clark2020ELECTRA}, which aims to improve computational efficiency. The models are pre-trained on the German part of OSCAR \citep{ortizsuarez:hal-02148693}, a set of monolingual corpora based on Common Crawl \citep{wenzek-etal-2020-ccnet}, which is a repository for multilingual web crawl data. Additionally, the pre-training data includes, in smaller parts, dumps from Wikipedia and text from a range of domains such as court decisions, movie subtitles, speeches, or books. GottBERT \citep{scheible2020gottbert} presents another BERT-flavoured German language model. It is trained on the German part of OSCAR. It uses the RoBERTa pre-training setup \citep{liu2019roberta}, aiming to optimize the training setup (e.g., hyperparameter values) of the original BERT setup \cite{devlin-etal-2019-bert}. 

\paragraph{Multilingual models} We choose competitive multilingual models based on the MTEB leaderboard. This includes two pre-trained English models (MiniLM-L12-v2-ml, MPNet-base-v2-ml) fine-tuned using multilingual knowledge distillation \cite{reimers-gurevych-2020-making}. Another model, USE-CMLM-ml, uses an adapted masked language modeling technique for training \cite{yang-etal-2021-universal}. SRoBERTa-cross, a Hugging Face community model, is based on the small variant of XLM-RoBERTa, a RoBERTa model trained on data in over 100 languages from Common Crawl, and then fine-tuned for German-English sentence similarity. We also use Sentence-T5 (ST5) encoders \cite{ni-etal-2022-sentence}. These models are based on the multilingual general-purpose T5 encoder-decoder model \cite{raffel202-t5} and fine-tuned on a large English dataset for sentence similarity. All of these models use training techniques designed to improve short text representations. Therefore, we also select XLM-RoBERTa-large as a more general-purpose multilingual model. It uses the masking technique from BERT \cite{devlin-etal-2019-bert} for pre-training.

\begin{figure*}[!t]
  \centering
  \resizebox{1.0\textwidth}{!}{
  \includesvg{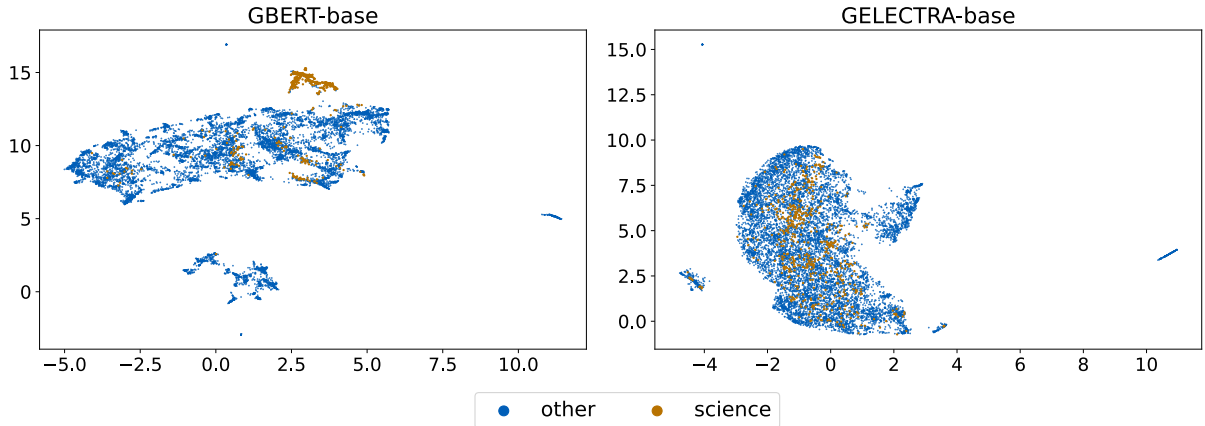}
  }
  \caption{
  Word embeddings of all texts from the TenkGnadClusteringP2P dataset reduced to two dimensions with UMAP. Texts belonging to the news category \textit{Wissenschaft} (science) are highlighted by color and size.
}
\label{fig:embed-cloud-gmodels}
\end{figure*}

\subsection{Clustering}
\paragraph{Algorithms} Like MTEB, we use Minibatch k-Means \cite{sklearn_api} and V-measure as an evaluation metric. Additionally, we perform analyses for Agglomerative Clustering \cite{sklearn_api}, DBSTREAM \cite{montiel2021river}, and HDBSCAN \cite{mcinnes2017hdbscan}. We select these algorithms based on their ease of use (e.g., pip-ready package), popularity, and abilities: Agglomerative Clustering is distance-based, similar to Minibatch k-Means. However, it may be more suited for modeling clusters of varying shapes and sizes. HDBSCAN, a density-based clustering algorithm, is used per default by the increasingly popular BERTopic. Lastly, DBSTREAM is a density-based algorithm for evolving data streams. In principle, DBSTREAM could cluster documents for real-time analysis (e.g., news monitoring). We use default parameters for all algorithms, focusing on the out-of-the-box performance. DBSTREAM does not support setting the number of clusters. In all other cases, we provide clustering models with this information (similar to MTEB).

\paragraph{Dimensionality reduction} We also experiment with dimensionality reduction to cluster lower dimensional data, motivated by the curse of dimensionality \cite{beyer1999, aggarwal2001}. We reduce the embedding vectors to two dimensions for every language model using PCA \cite{sklearn_api}, a standard technique and UMAP \cite{mcinnes2018umap-software}, which BERTopic suggests.

\subsection{Adaptive Pre-training}
We conduct experiments with adaptive pre-training separately for each dataset described in Subsection~\ref{subsec:ger-add}. In general, we assume an application scenario, where clustering is used to unsupervisingly analyze an ongoing text-based information feed, e.g., Twitter. Here, the focus lies not on the extrapolation capability towards new unseen data but on a consistent clustering of the entire text body. Thus, we use the evaluation data simultaneously as training data. This setup allows us to draw real-world conclusions, whether continued pre-training as an additional step before clustering can reliably improve clustering outcomes. We experiment with two pre-training techniques and evaluate them through V-measure. Firstly, we train with the general-purpose WWM technique (similar to GBERT). Given the relatively small training datasets, we follow the parameter setup for task-adaptive pre-training suggested in \citet{gururangan-etal-2020-dont}. We also experiment with the Transformers and Sequential Denoising Auto-Encoder (TSDAE) method \citep{wang-etal-2021-tsdae-using}, a state-of-the-art unsupervised training method for improving sentence embeddings, and we use the suggested parameter setup by the authors.


We use GBERT for these experiments. GBERT perform competitively (as discussed in Section~\ref{sec:res-dis}), and we are interested in the potential improvements for such strong models. What is more, GBERT uses the general-purpose WWM for pre-training, allowing it to evaluate the effect of the more task-specific TSDAE training. Generally, we want to provide some intuition for the potential use of these training methods specifically for clustering.

\begin{table*}[!t]
\centering
\resizebox{0.8\textwidth}{!}{
\begin{tabular}{l|rrrrrr|r}
                  & \multicolumn{2}{c}{\textbf{Blurbs}} & \multicolumn{2}{c}{\textbf{TenkGnad}} & \multicolumn{2}{c|}{\textbf{Reddit}} &                \\
\textbf{Model}    & \textbf{S2S}     & \textbf{P2P}     & \textbf{S2S}      & \textbf{P2P}      & \textbf{S2S}      & \textbf{P2P}     & \textbf{Avg.}  \\ \hline
GBERT-base        & 11.27            & 35.36            & 24.23             & 37.16             & 28.57             & 35.30            & 28.65          \\
GBERT-large       & 13.38            & 39.30            & \textbf{34.97}    & 41.69             & 34.47             & 44.61            & 34.74          \\
GELECTRA-base     & 7.74             & 10.06            & 4.11              & 9.02              & 6.59              & 7.73             & 7.54           \\
GELECTRA-large    & 7.57             & 13.96            & 3.91              & 11.49             & 7.59              & 10.54            & 9.18           \\
GottBERT          & 8.37             & 34.49            & 9.34              & 33.66             & 16.07             & 19.46            & 20.23          \\
MiniLM-L12-v2-ml  & 14.33            & 32.46            & 22.26             & 36.13             & 33.34             & 44.59            & 30.52          \\
MPNet-base-v2-ml  & 15.81            & 34.38            & 22.00             & 35.96             & 36.39             & 48.43            & 32.16          \\
SRoBERTa-cross    & 12.69            & 30.82            & 10.94             & 23.50              & 27.98             & 33.01            & 23.16          \\
USE-CMLM-ml       & 15.24            & 29.63            & 25.64             & 37.10             & 33.62             & 49.70            & 31.82          \\
ST5-base          & 11.57            & 30.59            & 18.11             & \textbf{44.88}    & 31.99             & 45.80            & 30.49          \\
ST5-xxl           & \textbf{15.94}   & \textbf{39.91}   & 19.69             & 43.43             & \textbf{38.54}    & \textbf{55.90}   & \textbf{35.57} \\
XLM-RoBERTa-large & 7.29             & 29.84            & 6.16              & 32.46             & 10.19             & 23.50            & 18.24         
\end{tabular}
}
\caption{
V-measure scores for the benchmark results of all evaluated models using the Minibatch k-Means algorithm. Results are multiplied by 100 and rounded to two decimals. \textbf{Bold} numbers indicate best column-wise result.
}
\label{tab:kmeans-results}
\end{table*}

\begin{table*}[!t]
\centering
\resizebox{0.8\textwidth}{!}{
\begin{tabular}{l|rrrrrr|r}
                                 & \multicolumn{2}{c}{\textbf{Blurbs}} & \multicolumn{2}{c}{\textbf{TenKGnad}} & \multicolumn{2}{c|}{\textbf{Reddit}} & \multicolumn{1}{l}{}          \\
\textbf{Algorithm}               & \textbf{S2S}     & \textbf{P2P}     & \textbf{S2S}      & \textbf{P2P}      & \textbf{S2S}      & \textbf{P2P}     & \textbf{Avg.}                 \\ \hline
Minibatch   k-Means              & 11.77            & 30.07            & 16.78             & 32.21             & 25.44             & 34.88            & 25.19                         \\
\textit{PCA-reduced embeddings}  & \textit{9.40}    & \textit{23.50}   & \textit{11.41}    & \textit{20.56}    & \textit{11.95}    & \textit{16.10}   & \textit{15.49}                \\
\textit{UMAP-reduced embeddings} & \textit{12.65}   & \textit{29.58}   & \textit{21.76}    & \textit{39.73}    & \textit{28.56}    & \textit{41.28}   & {\ul \textit{28.93}}          \\ \hline
Agglomerative Clustering         & 12.45            & 30.40            & 17.25             & 34.18             & 25.74             & 35.86            & 25.98                         \\
\textit{PCA-reduced embeddings}  & \textit{9.33}    & \textit{23.30}   & \textit{11.03}    & \textit{20.24}    & \textit{11.89}    & \textit{16.67}   & \textit{15.41}                \\
\textit{UMAP-reduced embeddings} & \textit{12.72}   & \textit{32.88}   & \textit{21.45}    & \textit{39.94}    & \textit{28.65}    & \textit{41.50}   & {\ul \textit{\textbf{29.52}}} \\ \hline
HDBSCAN                          & \multicolumn{6}{c|}{n/a}                                                                                           & \multicolumn{1}{c}{n/a}       \\
\textit{PCA-reduced embeddings}  & \textit{9.68}    & \textit{13.12}   & \textit{7.08}     & \textit{10.60}    & \textit{14.58}    & \textit{16.83}   & \textit{11.98}                \\
\textit{UMAP-reduced embeddings} & \textit{14.98}   & \textit{22.51}   & \textit{14.46}    & \textit{27.95}    & \textit{24.19}    & \textit{30.61}   & {\ul \textit{22.45}}          \\ \hline
DBSTREAM                         & \multicolumn{6}{c|}{n/a}                                                                                           & \multicolumn{1}{c}{n/a}       \\
\textit{PCA-reduced embeddings}  & \textit{6.41}    & \textit{14.19}   & \textit{7.79}     & \textit{12.46}    & \textit{8.38}     & \textit{10.68}   & \textit{9.99}                 \\
\textit{UMAP-reduced embeddings} & \textit{12.93}   & \textit{31.41}   & \textit{22.56}    & \textit{38.27}    & \textit{28.61}    & \textit{36.59}   & {\ul \textit{28.40}}         
\end{tabular}
}
\caption{
Average V-measure score of all evaluated models using different clustering algorithms and reduced embeddings as input (in \textit{italic}). Results are multiplied by 100 and rounded to two decimals. The \textbf{bold} number indicates the best overall result and \underline{underlined} results the best result per clustering algorithm.
}
\label{tab:algs-results}
\end{table*}

\section{Results and Discussion}
\label{sec:res-dis}
\subsection{Baseline: Minibatch k-Means}
\paragraph{Monolingual models} GBERT models perform better than the other monolingual GELECTRA and GottBERT models, as shown in Table~\ref{tab:kmeans-results}. GBERT-large ranks second best of all evaluated models. All models perform relatively better on P2P datasets compared to the S2S counterparts. This is an intuitive result, considering that models are presented with more information in these tasks. The weak performance of the GELECTRA models is surprising, given the strong results on downstream tasks reported in \citet{chan-etal-2020-germans}. Figure~\ref{fig:embed-cloud-gmodels} provides some visual intuition for this performance lack. For the news articles from the TenKGnadClusteringP2P dataset, GELECTRA produces more evenly-spread embeddings than GBERT, and embeddings belonging to the same topic (such as science) tend to be more spread. GottBERT lies in the middle between the GELECTRA and GBERT models. The gap to GBERT models is likely caused by GottBERT's smaller and less diverse training data. As discussed in Subsection~\ref{subsec:eval-models}, GBERT models contain training data that is more similar to the characteristics of the evaluation datasets (e.g., books and shorter text sequences such as movie subtitles).

\paragraph{Multilingual models} Apart from SRoBERTa-cross and XLM-RoBERTA-large, multilingual models perform competitively with scores close to the monolignual GBERT-base and GBERT-large. ST5-xxl is the best-performing model overall, scoring best on five out of six datasets (Table~\ref{tab:kmeans-results}). Moreover, the scaled ST5-xxl model (4.8B parameters) shows clear performance gains compared to its base variant (ST5-base, 110M parameters). ST5 models' fine-tuning data includes Reddit data, which may explain the strong performance on German Reddit datasets, i.e., a robust cross-lingual transfer.
The relatively weak results for XLM-RoBERTa-large are likely caused by less diverse training data and more general pre-training compared to the other multilingual models. The performance of SRoBERTa-cross, based on the smaller version of the XLM-RoBERTA-large model and fine-tuned for sentence similarity, also points in this direction, performing better on five out of six datasets than XLM-RoBERTA-large.

\subsection{Beyond k-Means}
Table~\ref{tab:algs-results} reports the average V-measure score of all evaluated models for different clustering algorithms. We do not report results for DBSTREAM and HDBSCAN for non-reduced embeddings, as we observed very poor computational performance for high-dimensional data during our experiments. In any case, results for Minibatch k-Means and Agglomerative Clustering suggest possibly better performance with reduced embeddings: Using UMAP-reduced embeddings improves the Minibatch k-Means and Agglomerative Clustering scores, on average, by around +13-15\% (compared to not using any reduction at all). What is more, clustering in low dimensions may benefit the explainability of models as it allows to visually analyze results (e.g., Figure~\ref{fig:embed-cloud-gmodels}). However, this does not hold for clustering with PCA-reduced embeddings, which show the worst results by far. This limits its use for text-based clustering.

Overall, and based on the results for clustering with UMAP-reduced embeddings, Minibatch k-Means and Agglomerative Clustering perform very similarly. DBSTREAM performs slightly worse on average, caused by relatively weaker results for P2P datasets.\footnote{Shortly before the final submission, we found a small bug in the DBSTREAM implementation we used: \url{https://github.com/online-ml/river/issues/1265}. However, we do not expect a significant influence on the overall results.} HDBSCAN performs worst on five out of six datasets. We suspect the weaker results for HDBSCAN are caused by its sensitivity to classifying data points as noise. In our experiments, we observe that in some cases, more than 30\% of the data is labeled as noise. A different configuration for HDBSCAN would likely improve results. However, the usefulness of such sensitive algorithms may also depend on the use case (e.g., whether any text data is considered as noise).

\begin{figure*}[!t]
  \centering
  \resizebox{1\textwidth}{!}{
  \includesvg{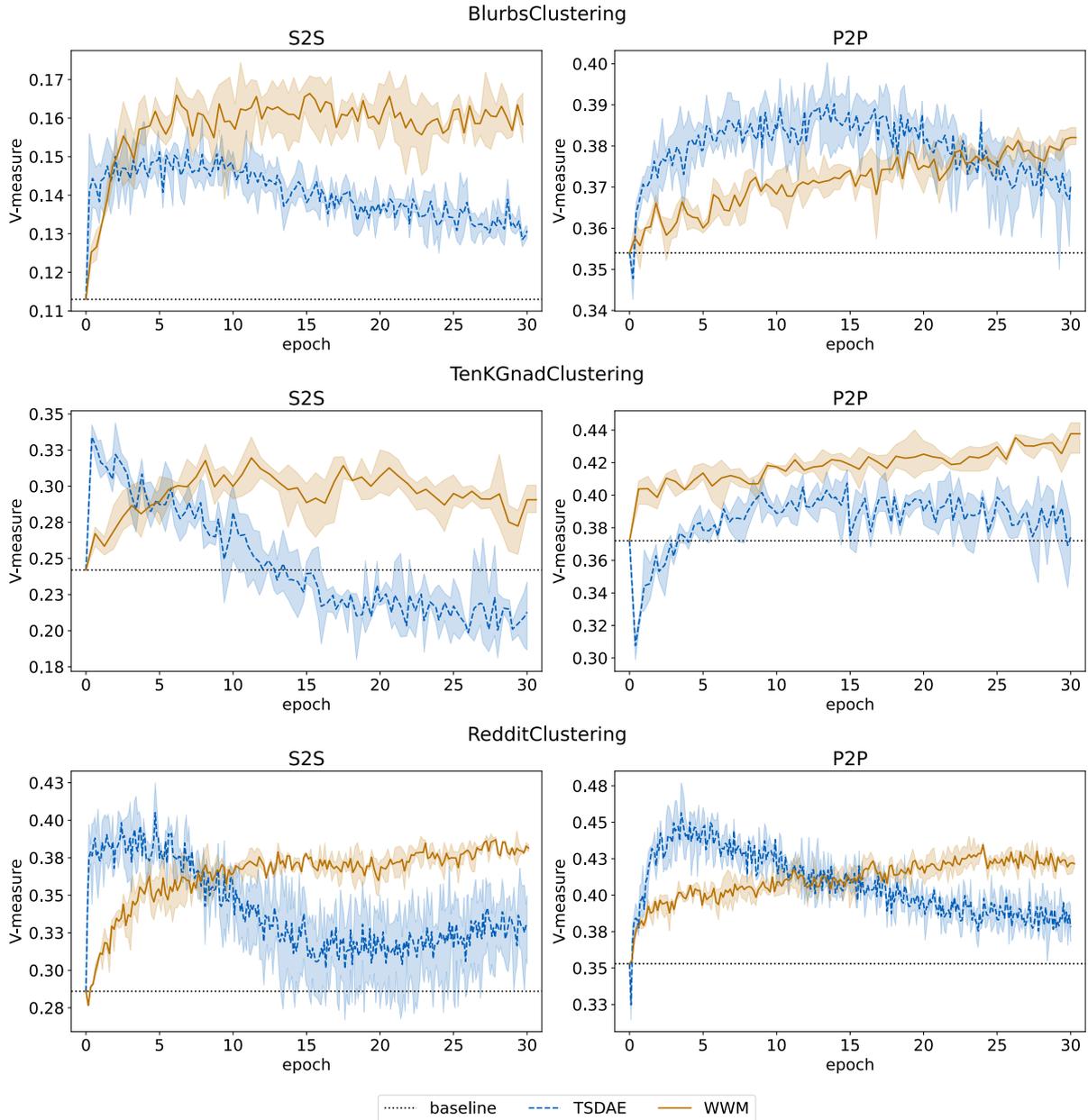}
  }
  \caption{
Change of the V-measure score with continued pre-training for GBERT-base comparing WWM and TSDAE pre-training methods for Minibatch k-Means clustering. Lines represent the average of three model runs with different seeds, and filled areas indicate minimum and maximum V-measure scores. \textit{baseline} indicates results without additional pre-training.
}
\label{fig:gbert-base-ft}
\end{figure*}

\subsection{Adaptive Pre-training with GBERT}
The evaluation for GBERT-base and adaptive pre-training with WWM and TSDAE, reported in Figure~\ref{fig:gbert-base-ft}, shows clear performance improvements for all benchmark datasets.\footnote{We provide exact numbers in Table~\ref{tab:ft-add-results} in the appendix.} The improvements on the S2S datasets are considerably more significant for both pre-training methods: After around one epoch of TSDAE training, V-measures improve by around +31\% on average. After around 10 epochs of WWM training, V-measures improve by around +30\% on average. For P2P datasets, the improvements are relatively smaller, and models profit from more extended WWM training (average improvement of around +15\% after 30 epochs). The benefit of TSDAE training seems less clear, requiring more extended training to compensate for initial performance drops and different training times to reach maximum improvements. Overall, and based on the results for 10 epochs of WWM training (Table~\ref{tab:ft-add-results}), GBERT-base converges to the performance of the larger GBERT-large (33.70 vs. 34.74) and ranks second best out of all models for all S2S dataset. We suspect the more significant improvements on smaller text sequences (S2S) are likely caused by the fact that GBERT models are generally pre-trained on much longer sequences (maximum of 512 subword tokens per sample), considering that S2S samples are, on average, only around up to 50 characters long (Table~\ref{tab:datasets}). This may also explain why the improvements for P2P datasets are relatively minor, as these texts more closely resemble the pre-training data of the unadopted GBERT (in terms of text length).

We performed similar experiments for GBERT-large, as shown in Figure~\ref{fig:gbert-large-ft} in the appendix. In most cases, and for both pre-training methods, the performance decreases significantly (e.g., TenKGnadClusteringS2S with TSDAE) or stays relatively unchanged (e.g., TenKGnadClusteringS2S with WWM). The training stability seems low as different training with different seeds may result in different performances (see also Table~\ref{tab:ft-add-results}). We suspect the relatively low batch sizes (256 for WWM and eight for TSDAE compared to 2,048 for GBERT-large's previous pre-training) lead to these training instabilities, as parameter updates are too aggressive. The parameter setups we used are based on experiments with BERT models similar to GBERT-base in terms of parameters \cite{gururangan-etal-2020-dont,wang-etal-2021-tsdae-using}. Our results suggest that these setups are unsuitable for larger models.

\section{Conclusion}
This work introduces German benchmark datasets for the evaluation of embedding-based clustering, building on the monolingual clustering benchmark from Massive Text Embedding Benchmark (MTEB). We introduce six datasets from three sources (blurbs, news articles, and Reddit). Additionally, we evaluate the out-of-the-box performance of different clustering algorithms and show that UMAP-reduced embeddings improve clustering outcomes and simplify the visual analysis simultaneously. 
\\\indent In total, we evaluate 12 language models. Results are mixed as there are both strong (GBERT-large, ST5-xxl) and weak (GELECTRA-base, XLM-RoBERTa-large) monolingual and multilingual models (Table~\ref{tab:kmeans-results}). The selected models cover a wide range of different pre-training data, model sizes, and pre-training methods. A thorough investigation of how these factors influence clustering outcomes could build on this work.
\\\indent Lastly, we experiment with adaptive pre-training for GBERT models. We show that for GBERT-base, TSDAE and WWM pre-training drastically improves the performance for short texts and relatively modestly for longer texts. Results for the larger GBERT model are inconsistent and only show improvements in one case, which we suspect is caused by a too aggressive hyperparameter configuration. This leaves room for future experimentation, which would ideally include larger datasets.

\section*{Limitations}
\paragraph{Diversity of datasets} Compared to the MTEB clustering benchmark, our proposed German benchmark is less diverse. For instance, it does not contain formal writing (e.g., scientific papers). Moreover, the proposed datasets are relatively small with a maximum split size of around 26k samples (Table~\ref{tab:datasets}). Real-world applications may involve larger data (possibly hundreds of thousands of data samples) with a high degree of semantic variability (e.g., hundreds of topics), forcing models to perform extremely fine-grained clustering.

\paragraph{Pre-training experiments} Given the relatively small training datasets, our experiments do not allow us to conclude the possible benefits of more extensive data. In the case of larger available data, longer pre-training might be beneficial. Furthermore, the experiments focus on monolingual BERT-based model architecture. Benefits of continued pre-training may differ for, e.g., multilingual models or pre-trained models with smaller pre-training datasets (such as GottBERT).

\paragraph{Beneficial model properties} As discussed in Section~\ref{sec:res-dis}, some models perform very differently, although trained on similar data. From a practical point of view, a more thorough analysis of performance-increasing factors would be helpful (i.e., model size and architecture, pre-training method, and training data). Moreover, it would also be interesting to better understand how models assess the similarity of text. This could affect how well models are suited for specific clustering tasks (e.g., how models deal with words with specific grammatical functions or unseen words).

\paragraph{Large language models} The rise of generative large language models (LLMs), such as GPT-4 \cite{openai2023gpt4}, and primarily open-source models, such as LLaMA \cite{touvron2023llama}, are not represented in this work. While the benefit of generative models for NLU may not yet be fully understood, preliminary work suggests strong performance (e.g., \citealp{neelakantan2022text,muennighoff2022sgpt}). However, this work focuses on well-established models and training techniques that can be easily used with decent resources (e.g., a single GPU) and thus benefit the open-source community the most.

\section*{Ethics Statement}
\label{sec:eth-stat}
We acknowledge the ACL Code of Ethics\footnote{\url{https://www.aclweb.org/portal/content/acl-code-ethics}} as an essential instrument in ensuring that research in computer science serves the public good. In the context of this work, we want to discuss our approach to share user-owned social media data responsibly. Social media has become an integral part of everyday life and an important data source in many research fields. For instance, social media can be used to address information voids during health emergencies \cite{boender2023}. Consequently, the use of social media data in NLU research and applications has increased, rendering the inclusion of such data in this benchmark (i.e., Reddit) essential.
\\\indent We use Reddit data for this benchmark for comparability (to MTEB) and because Reddit has open API access (meaning that any interested user can reproduce the published results). In fact, Reddit data is also available in large amounts without registration to the Reddit services: The Pushshift dataset \cite{Baumgartner_Zannettou_Keegan_Squire_Blackburn_2020} has been collecting any public Reddit data for over a decade, providing the collected data to anyone and without any form of authentication. We believe this is a problematic approach as the data is distributed without requiring parties to accept the Reddit API terms of use. Specifically, and as per the current terms, Reddit data is owned by Reddit users (and not the platform itself), allowing users to delete accounts and content.\footnote{Reddit recently updated the API terms, which became effective on June 19, 2023 (\url{https://www.redditinc.com/policies/data-api-terms}). The updated terms define a less permissive use of Reddit data for artificial intelligence applications, and interested researchers should carefully consider these terms. This work was performed under the old, more permissive API terms.} Specifically, the General Data Protection Regulation (GDPR), applicable to member states of the European Union (EU), mandates the right of the deletion of personal data ("the right to be forgotten").\footnote{\url{https://gdpr.eu/right-to-be-forgotten}} Deliberately sharing user-owned data to anonymous parties makes it practically impossible for users to invoke their rights. Instead, data should only be obtained through the official Reddit API, which can be used to obtain and update Reddit data. Therefore, we do not disclose the raw data and instead, only share data identifiers and advise interested researchers to use the official channels.


\bibliography{anthology,custom}
\bibliographystyle{acl_natbib}

\clearpage
\appendix

\onecolumn
\section{Additional Results}

\begin{figure}[h!]
\centering
\resizebox{1.0\textwidth}{!}{
\includesvg{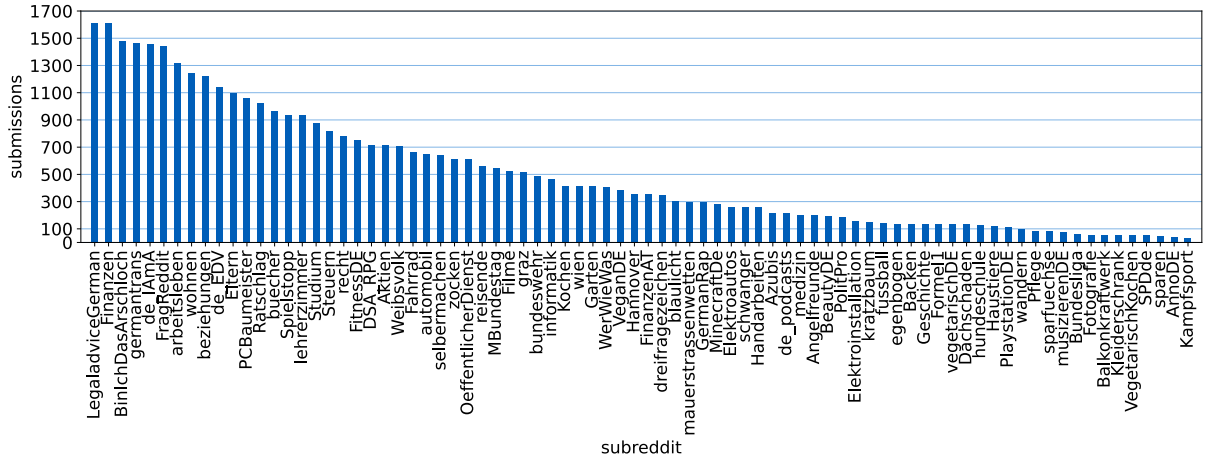}
}
\caption{
Distribution of German Reddit dataset used for RedditClusteringS2S and RedditClusteringP2P.
}
\label{fig:reddit-bars}
\end{figure}

\begin{table}[h!]
\centering
\resizebox{1.0\textwidth}{!}{
\begin{tabular}{l|l}
\textbf{Model}    & \textbf{Hugging Face Repository}                                                   \\ \hline
GBERT-base        & \url{https://huggingface.co/deepset/gbert-base}                                          \\
GBERT-large       & \url{https://huggingface.co/deepset/gbert-large}                                         \\
GELECTRA-base     & \url{https://huggingface.co/deepset/gelectra-base}                                       \\
GELECTRA-large    & \url{https://huggingface.co/deepset/gelectra-large}                                      \\
GottBERT          & \url{https://huggingface.co/uklfr/gottbert-base}                                         \\
MiniLM-L12-v2-ml  & \url{https://huggingface.co/sentence-transformers/paraphrase-multilingual-MiniLM-L12-v2} \\
MPNet-base-v2-ml  & \url{https://huggingface.co/sentence-transformers/paraphrase-multilingual-mpnet-base-v2} \\
SRoBERTa-cross    & \url{https://huggingface.co/T-Systems-onsite/cross-en-de-roberta-sentence-transformer}   \\
USE-CMLM-ml       & \url{https://huggingface.co/sentence-transformers/use-cmlm-multilingual}                 \\
ST5-base          & \url{https://huggingface.co/sentence-transformers/sentence-t5-base}                      \\
ST5-xxl           & \url{https://huggingface.co/sentence-transformers/sentence-t5-xxl}                       \\
XLM-RoBERTa-large & \url{https://huggingface.co/xlm-roberta-large}                                          
\end{tabular}
}
\caption{
Hugging Face repositories for all evaluated language models.
}
\label{tab:hf-model}
\end{table}

\begin{figure}
  \centering
  \resizebox{1.0\textwidth}{!}{
  \includesvg{plots/gbert-large-ft.svg}
  }
  \caption{
  Change of the V-measure score with continued pre-training for GBERT-large comparing WWM and TSDAE pre-training methods for Minibatch k-Means clustering. Lines represent the average of three model runs with different seeds, and filled areas indicate minimum and maximum V-measure scores. \textit{baseline} indicates results without additional pre-training.
}
\label{fig:gbert-large-ft}
\end{figure}

\clearpage
\begin{sidewaystable}
\centering
\resizebox{1.0\textwidth}{!}{
\begin{tabular}{cl|rrrrrr|rrrrrr|rrrrrr|rr}
\multicolumn{1}{l}{}            &                       & \multicolumn{6}{c|}{\textbf{Blurbs}}                                           & \multicolumn{6}{c|}{\textbf{TenKGnad}}                                         & \multicolumn{6}{c|}{\textbf{Reddit}}                                           & \multicolumn{1}{l}{} & \multicolumn{1}{l}{} \\
\multicolumn{1}{l}{}            &                       & \multicolumn{3}{c|}{\textbf{S2S}}          & \multicolumn{3}{c|}{\textbf{P2P}} & \multicolumn{3}{c|}{\textbf{S2S}}          & \multicolumn{3}{c|}{\textbf{P2P}} & \multicolumn{3}{c|}{\textbf{S2S}}          & \multicolumn{3}{c|}{\textbf{P2P}} & \multicolumn{1}{c}{} & \multicolumn{1}{c}{} \\
\multicolumn{1}{l}{\textbf{}}   & \textbf{Method} & SD   & Avg.  & \multicolumn{1}{r|}{$\Delta$}      & SD        & Avg.      & $\Delta$         & SD   & Avg.  & \multicolumn{1}{r|}{D}      & SD        & Avg.      & $\Delta$         & SD    & Avg.  & \multicolumn{1}{r|}{$\Delta$}     & SD        & Avg.      & $\Delta$         & \textbf{Avg.}        & \textbf{$\Delta$}           \\ \cline{2-22} 
\multirow{3}{*}{\textbf{base}}  & TSDAE (1)             & 0.46 & 14.37 & \multicolumn{1}{r|}{+3.10}  & 0.29      & 37.08     & +1.72     & 1.40 & 32.87 & \multicolumn{1}{r|}{+8.64}  & 0.79      & 34.73     & -2.43     & 0.42  & 37.29 & \multicolumn{1}{r|}{+8.72} & 0.33      & 39.03     & +3.73      & 32.56                & +3.91                \\
                                & WWM (10)              & 0.80 & 15.71 & \multicolumn{1}{r|}{+4.44}  & 0.41      & 36.83     & +1.47     & 1.12 & 30.81 & \multicolumn{1}{r|}{+6.58}  & 0.27      & 41.47     & +4.31      & 0.53  & 36.48 & \multicolumn{1}{r|}{+7.91} & 0.87      & 40.90     & +5.60      & 33.70                & +5.05                \\
                                & WWM (30)              & \multicolumn{3}{c|}{n/a}                   & 0.18      & 38.20     & +2.84     & \multicolumn{3}{c|}{n/a}                   & 0.85      & 43.77     & +6.61      & \multicolumn{3}{c|}{n/a}                   & 0.25      & 42.16     & +6.86      & 41.38                & +12.73               \\ \cline{2-22} 
\multirow{2}{*}{\textbf{large}} & TSDAE (1)             & 0.87 & 2.57  & \multicolumn{1}{r|}{-10.81} & 1.35      & 38.48     & -0.82     & 0.36 & 0.82  & \multicolumn{1}{r|}{-34.15} & 7.47      & 35.60     & -6.09     & 17.63 & 25.90 & \multicolumn{1}{r|}{-8.57} & 8.65      & 48.89     & +4.28      & 25.38                & -9.36                \\
                                & WWM (10)              & 6.98 & 11.89 & \multicolumn{1}{r|}{-1.49}  & 0.57      & 36.70     & -2.60     & 1.60 & 32.07 & \multicolumn{1}{r|}{-2.90}  & 0.58      & 42.98     & +1.29      & 17.99 & 26.04 & \multicolumn{1}{r|}{-8.43} & 1.53      & 40.74     & -3.87     & 31.74                & -3.00               
\end{tabular}
}
\caption{V-measure scores for GBERT-base (\textbf{base}) and GBERT-large (\textbf{large}) with continued pre-training for TSDAE and WWM methods. Evaluated is based on Minibatch k-Means clustering, and reported V-measure scores are multiplied by 100. Results are multiplied by 100 and rounded to two decimals. Numbers in brackets denote the number of epochs. \textit{Avg.} represents the average of three training runs with different seeds and \textit{SD} the standard deviation of V-measures. $\Delta$ reports the absolute difference to the baseline, i.e., the model results without additional pre-training.}
\label{tab:ft-add-results}
\end{sidewaystable}

\end{document}